\documentclass[10pt,twocolumn,letterpaper]{article}

\usepackage{cvpr}      

\usepackage{graphicx}
\usepackage{amsmath}
\usepackage{amssymb}
\usepackage{booktabs}
\usepackage{multirow}
\usepackage{color}
\usepackage{xcolor}
\definecolor{wrong}{rgb}{.8,.349,.1}
\definecolor{right}{rgb}{.3,.7,.1}
\usepackage{newfloat}
\usepackage{listings}
\usepackage{amsmath}
\usepackage{amssymb}

\usepackage[pagebackref,breaklinks,colorlinks]{hyperref}

\usepackage[capitalize]{cleveref}
\crefname{section}{Sec.}{Secs.}
\Crefname{section}{Section}{Sections}
\Crefname{table}{Table}{Tables}
\crefname{table}{Tab.}{Tabs.}

\def\cvprPaperID{841} 
\def\confName{CVPR}
\def\confYear{2023}

\begin{document}

\title{Bidirectional Copy-Paste for Semi-Supervised Medical Image Segmentation}

\author{Yunhao Bai\textsuperscript{1} \quad Duowen Chen\textsuperscript{1} \quad Qingli Li\textsuperscript{1} \quad Wei Shen\textsuperscript{2} \quad Yan Wang\textsuperscript{1}\footnotemark[1]\\
\normalsize{\textsuperscript{1}Shanghai Key Laboratory of Multidimensional Information Processing, East China Normal University}\\
\normalsize{\textsuperscript{2}MoE Key Lab of Artificial Intelligence, AI Institute, Shanghai Jiao Tong University}\\
{\tt\small \{yhbai@stu.,~duowen\_chen@stu.,~qlli@cs.,~ywang@cee.\}ecnu.edu.cn, wei.shen@sjtu.edu.cn}\\
}
\maketitle
\renewcommand*{\thefootnote}{\fnsymbol{footnote}}
 \setcounter{footnote}{1}
 \footnotetext{Corresponding Author.}

\begin{abstract}\vspace{-0.6em}
In semi-supervised medical image segmentation, there exist empirical mismatch problems between labeled and unlabeled data distribution.
The knowledge learned from the labeled data may be largely discarded if treating labeled and unlabeled data \emph{separately} or {in an \emph{inconsistent} manner}.
We propose a straightforward method for alleviating the problem $-$ copy-pasting labeled and unlabeled data bidirectionally, in a simple Mean Teacher architecture. The method encourages unlabeled data to learn comprehensive common semantics from the labeled data in both \emph{inward} and \emph{outward} directions. More importantly, the consistent learning procedure for labeled and unlabeled data can largely reduce the empirical distribution gap. 
In detail, we copy-paste a random crop from a labeled image (foreground) onto an unlabeled image (background) and an unlabeled image (foreground) onto a labeled image (background), respectively. {The two mixed images are fed into a Student network and supervised by the mixed supervisory signals of pseudo-labels and ground-truth.} 
We reveal that the simple mechanism of copy-pasting bidirectionally between labeled and unlabeled data is good enough and the experiments show solid gains (\emph{e.g.}, over {21}\% Dice improvement on ACDC dataset with 5\% labeled data) compared with other state-of-the-arts on various semi-supervised medical image segmentation datasets. Code is available at \url{https://github.com/DeepMed-Lab-ECNU/BCP}\vspace{-0.8em}. 
\end{abstract}

\begin{figure}[t]
\includegraphics[width=0.7\columnwidth]{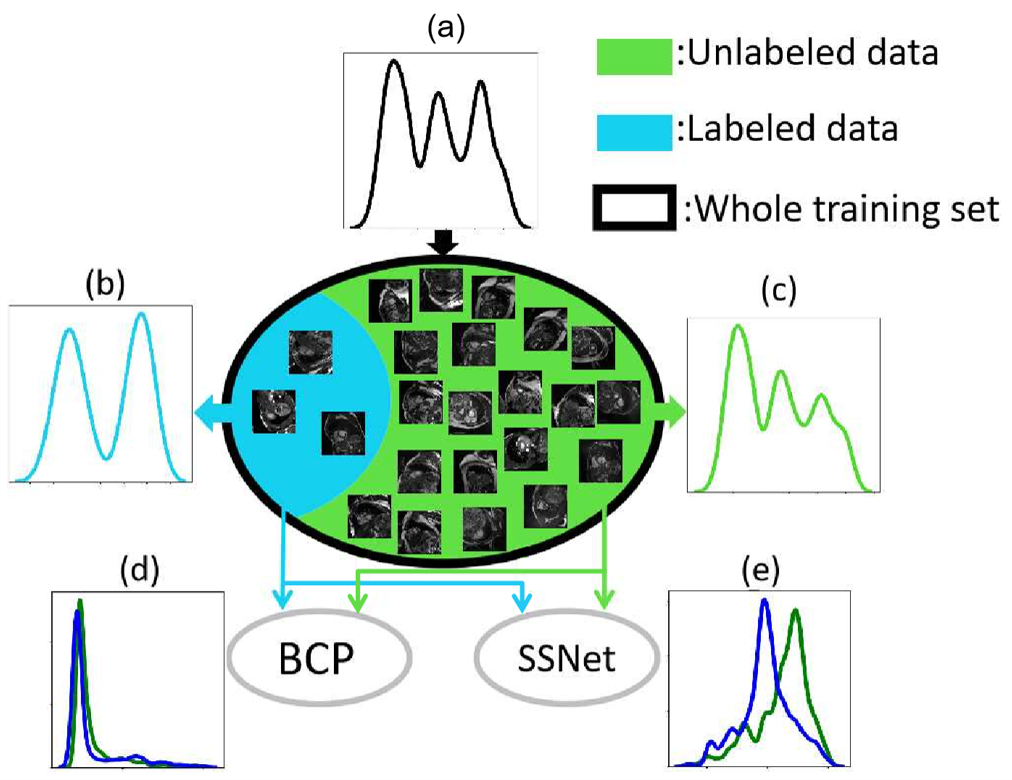}
\centering
\caption{Illustration of the mismatch problem under semi-supervised leaning setting. Assume the training set is drawn from a latent distribution in \textbf{(a)}. But the empirical distributions of small amount of labeled data and a large amount of unlabeled data are \textbf{(b)} and \textbf{(c)}, respectively. It's hard to use few labeled data to construct the precise distribution of the whole dataset. 
\textbf{(d)} By using our BCP, the empirical distributions of labeled and unlabeled features are aligned. \textbf{(e)} But other methods such as SSNet \cite{SSNet} or cross unlabeled data copy-paste cannot address the empirical distribution mismatch issue. All distributions are kernel density estimations of voxels belonging to myocardium class in ACDC \cite{ACDCdataset}.}
\label{fig:example}
\vspace{-1em}
\end{figure}

\section{Introduction}
\label{sec:intro}
Segmenting internal structures from medical images such as computed tomography (CT) or magnetic resonance imaging (MRI) is essential for many clinical applications \cite{Wang2019abdominal}. Various techniques based on supervised learning for medical image segmentation have been proposed \cite{Li2018hdenseunet,Dou2020unpaired,Zhao20213d}, which usually requires a large amount of labeled data. But, due to the tedious and expensive manual contouring process when labeling medical images, semi-supervised segmentation attracts more attention in recent years, and has become ubiquitous in the field of medical image analysis.

{Generally speaking, in semi-supervised medical image segmentation, the labeled and unlabeled data are drawn from the same distribution, (Fig.~\ref{fig:example} (a)). But in real-world scenario, it's hard to estimate the precise distribution from labeled data because they are few in number. Thus, there always exists \textbf{empirical distribution mismatch} between a large amount of unlabeled and a very small amount of labeled data \cite{SSL_ADA} (Fig.~\ref{fig:example}(b) and (c)). }Semi-supervised segmentation methods always attempt to train labeled and unlabeled data symmetrically, in a consistent manner. \emph{E.g.}, self-training \cite{Bai2017semi,Zhou2019semi} generates pseudo-labels to supervise unlabeled data in a pseudo-supervised manner. Mean Teacher based methods \cite{SimCVD} adopt consistency loss to ``supervise'' unlabeled data with strong augmentations, in analogy with supervising labeled data with ground-truth. DTC \cite{Luo2021DTC} proposed a dual-task-consistency framework, applicable to both labeled and unlabeled data. ContrastMask \cite{Wang2022contrastmask} applied dense contrastive learning on both labeled and unlabeled data. But most existing semi-supervised methods used labeled and unlabeled data under separate learning paradigms. Thus, it often leads to the discarding of massive knowledge learned from the labeled data {and the empirical distribution mismatch between labeled and unlabeled data} (Fig.~\ref{fig:example}(e)).

CutMix \cite{CutMix} is simple yet strong data processing method, also dubbed as Copy-Paste (CP), which has the potential to encourage unlabeled data to learn common semantics from the labeled data, since pixels in the same map share semantics to be closer \cite{Wang2022separated}. {In semi-supervised learning, forcing consistency between weak-strong augmentation pair of unlabeled data is widely used \cite{Perturbed, FreeMatch, SSL_SC, Pixel_CC}, and CP is usually used as a strong augmentation.}
But existing CP methods only consider CP cross unlabeled data \cite{Perturbed, MUM, StrongVariedPerturbations}, or simply copy crops from labeled data as foreground and paste to another data \cite{SimpleCP, UCC}. They neglect to design a consistent learning strategy for labeled and unlabeled data, which hampers its usage on reducing the distribution gap. {{Meanwhile, CP tries to enhance the generalization of networks by increasing unlabeled data diversity, but a high performance is hard to achieve since CutMixed image is only supervised by low-precision pseudo-labels. It's intuitive to use more accurate supervision to help networks segment degraded region cut by CP.}} 

\begin{figure}
\includegraphics[width=1.0\columnwidth]{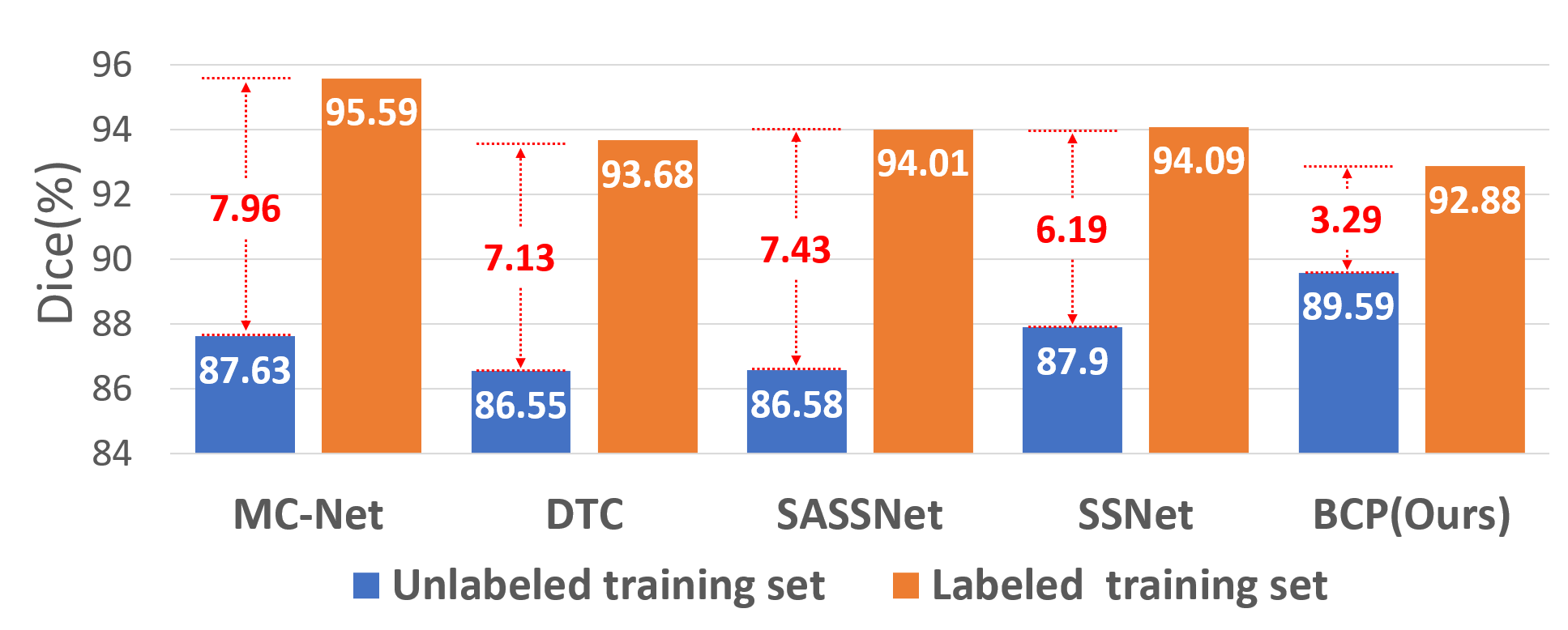}
\centering
\vspace{-0.6em}
\caption{Dice scores for unlabeled and labeled training data of different models on LA dataset \cite{LAdataset}. A much smaller performance gap is observed in our method.}
\label{fig:match}
\vspace{-0.9em}
\end{figure}

To alleviate the empirical mismatch problem between labeled and unlabeled data, a successful design is to encourage unlabeled data to learn comprehensive common semantics from the labeled data, and meanwhile, furthering the distribution alignment via a consistent learning strategy for labeled and unlabeled data. We achieve this by proposing a \textbf{surprisingly simple yet very effective} Bidirectional Copy-Paste (BCP) method, instantiated in the Mean Teacher framework. Concretely, to train the Student network, we augment our inputs by  copy-pasting random crops from a labeled image (foreground) onto an unlabeled image (background) and reversely, copy-pasting random crops from an unlabeled image (foreground) onto a labeled image (background). The Student network is supervised by the generated supervisory signal via bidirectional copy-pasting between the pseudo-labels of the unlabeled images from the Teacher network and the label maps of the labeled image. The two mixed images help the network to learn common semantics between the labeled and unlabeled data \textbf{bidirectionally and symmetrically}. We compute the Dice scores for labeled and unlabeled training set from LA dataset \cite{LAdataset} based on models trained by state-of-the-arts and our method, as shown in Fig.~\ref{fig:match}. {{Previous models which process labeled data and unlabeled data separately present strong performance gap between labeled and unlabeled data. \emph{E.g.}, MC-Net obtains 95.59\% Dice for labeled data but only 87.63\% for unlabeled data. It means previous models absorb knowledge from ground-truth well, but discard a lot when transferring to unlabeled data}}. Our method can largely decrease the gap between labeled and unlabeled data (Fig.~\ref{fig:example}(d)) in terms of their performances. It is also interesting to observe that Dice for labeled data of our BCP is lower than other methods, implying that BCP can mitigate the over-fitting problem to some extent.

We verify BCP in three popular datasets: LA \cite{LAdataset}, Pancreas-NIH \cite{NIHPancreas}, and ACDC \cite{ACDCdataset} datasets. Extensive experiments show our simple method outperforms all state-of-the-arts by a large margin, with even
over 21\% improvement in Dice on ACDC dataset with 5\% labeled data. Ablation study further shows the effectiveness of each proposed module. Note that compared with the baseline \emph{e.g.}, VNet or UNet, our method does not introduce new parameters for training, while remaining the same computational cost.


\section{Related Work}
\label{sec:relatedwork}

\subsection{Medical Image Segmentation}
Segmenting internal structures from medical images is essential for many clinical applications \cite{Wang2019abdominal}. Existing methods for medical image segmentation can be groups into two categories. The first category designed various 2D/3D segmentation network architectures \cite{Ronneberger2015unet,Cicek20163d,Milletari2016vnet,Zhou2018unet++,Li2018hdenseunet,Dou2020unpaired}. The second category leveraged medical prior knowledge to network training \cite{Tang2021weakly,Wang2021automatic,Xie2020recurrent,Wang2020deep}.

\begin{figure*}[t]
\includegraphics[width=0.7\linewidth]{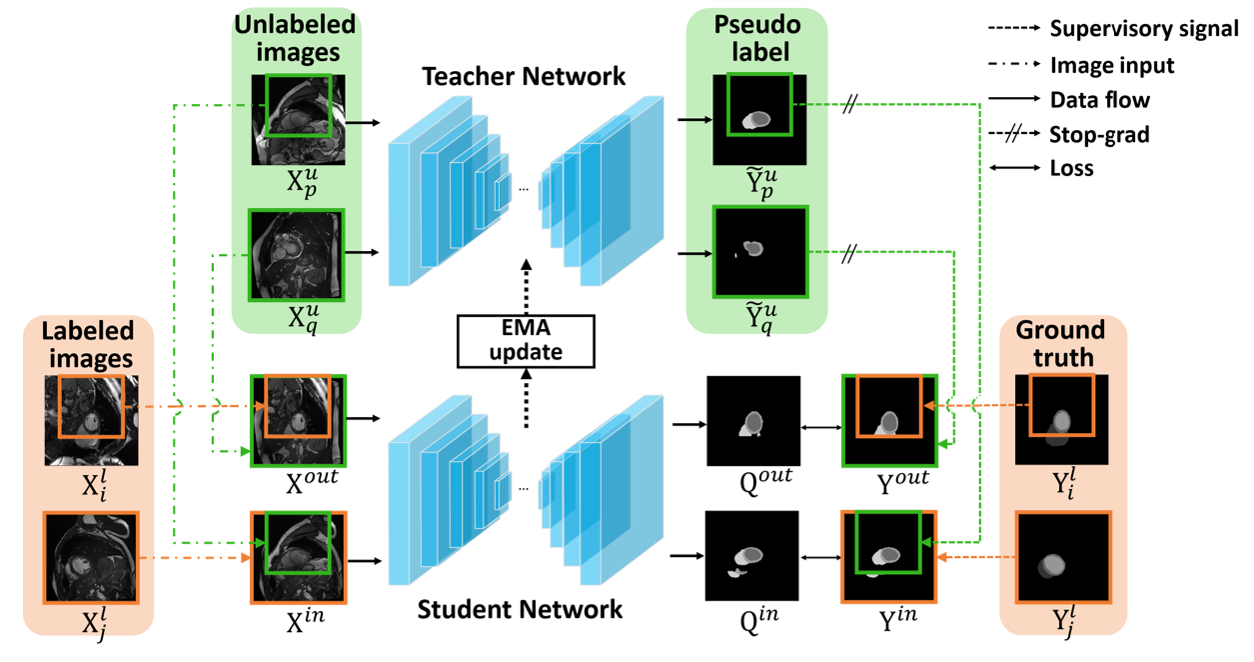}
\centering
\caption{Overview of our bidirectional copy-paste framework in Mean Teacher architecture, drawn with 2D inputs for better visualization. The inputs to Student network are generated by mixing two labeled and two unlabeled images in the proposed bidirectional copy-paste manner. Then, to provide the supervisory signal to the Student network, 
we combine the ground-truths and the pseudo-labels generated by the Teacher network into one supervisory signal via the same bidirectional copy-paste, to enable strong supervision from ground-truths help the weak supervision from pseudo-label.  
}
\label{fig:framework}
\vspace{-0.5em}
\end{figure*}

\subsection{Semi-supervised Medical Image Segmentation} 
Many efforts have been made in semi-supervised medical image segmentation. Entropy minimization (EM) and consistency regularization (CR) are the two widely-used loss functions. Meanwhile, many works extended Mean Teacher framework in different ways. SASSNet \cite{SASSNet} leveraged unlabeled data to enforce a geometric shape constraint on the segmentation output. DTC \cite{Luo2021DTC} proposed a dual-task-consistency framework by building task-level regularization explicitly. SimCVD \cite{SimCVD} modeled geometric structure and semantic information explicitly and constrain them between Teacher and Student networks. These methods used geometric constraints to supervise the output of the network. UA-MT \cite{UAMT} used uncertainty information to guide Student network learn from the meaningful and reliable targets of Teacher network gradually. \cite{Crosslevel} combined image-wise and patch-wise representations to explore more complex similarity cues, enforcing the output to be consistent given different input sizes. CoraNet \cite{CoraNet} proposed a model which can produce certain and uncertain regions, and Student network treats regions indicated from Teacher network with different weights. UMCT \cite{UMCT} used different perspectives of the network to predict the same image of different views. It utilized the predictions and the corresponding uncertainty to generate the pseudo-labels, which were used to supervise the prediction of unlabeled images. These methods have furthered the effectiveness for semi-supervised medical image segmentation. But, they ignored how to learn common semantics from labeled to unlabeled data. Treating labeled and unlabeled data separately often impedes knowledge transfer from labeled to unlabeled data.

\subsection{Copy-Paste} Copy-paste is a simple but strong data processing method for many tasks, \emph{e.g.}, instance segmentation \cite{SimpleCP,InstaBoost}, semantic segmentation \cite{UCC,GuidedMix-Net} and object detection \cite{ContextualCP}. Generally speaking, copy-paste means copying crops of one image and pasting them onto another image. Mixup \cite{Mixup} and CutMix \cite{CutMix} are classic works of mixing whole images and mixing image crops respectively. Many recent works extended them to address specific goals. GuidedMix-Net \cite{GuidedMix-Net} used mixup to generate higher-quality pseudo-labels by transferring the knowledge of labeled data to the unlabeled data. InstaBoost \cite{InstaBoost} and Contextual Copy-Paste \cite{ContextualCP} placed the cropped foreground onto another image elaborately according to the surrounding visual context. CP$^{2}$ \cite{CP2} proposed a pretraining method which copy-pastes a random crop from an image to another background image, it has been proved to be more suitable for downstream dense prediction tasks. \cite{SimpleCP} made a systematic study of copy-paste in instance segmentation. UCC \cite{UCC} copied the pixels belonging to the class which has a low confidence score as foreground during training to alleviate the distribution mismatch and class imbalance problems. Previous methods only considered copy-paste cross unlabeled data, or simply copied crops from labeled data as foreground and pasted to another data. They neglect to design a consistent learning strategy for labeled and unlabeled data. Thus, a large distribution gap is still inevitable. 

\section{Method}

Mathematically, we define the 3D volume of a medical image as $\textbf{X}\in\mathbb{R}^{W\times H\times L}$. The goal of  semi-supervised medical image segmentation is to predict the per-voxel label map $\widehat{\textbf{Y}}\in\{0,1, ..., K-1\}^{W\times H\times L}$, indicating where the background and the targets are in $\textbf{X}$. $K$ is the class number. Our training set $\mathcal{D}$ consists of $N$ labeled data and $M$ unlabeled data ($N\ll M$), expressed as two subsets: $\mathcal{D}=\mathcal{D}^l\cup \mathcal{D}^u$, where $\mathcal{D}^l = \{(\textbf{X}^l_i,\textbf{Y}^l_i)\}_{i=1}^N$ and $\mathcal{D}^u = \{\textbf{X}^u_i\}_{i=N+1}^{M+N}$.


The overall pipeline of the proposed bidirectional copy-paste method is shown in Fig.~\ref{fig:framework}, in the Mean Teacher architecture. We randomly pick two unlabeled images $(\textbf{X}^u_p, \textbf{X}^u_q)$, and two labeled images $(\textbf{X}^l_i, \textbf{X}^l_j)$ from the training set. Then we copy-paste a random crop from $\textbf{X}^l_i$ (the foreground) onto $\textbf{X}^u_q$ (the background) to generate the mixed image $\textbf{X}^{out}$, and from $\textbf{X}^u_p$ (the foreground) onto $\textbf{X}^l_j$ (the background) to generate another mixed image $\textbf{X}^{in}$. Unlabeled images are able to learn comprehensive common semantics from labeled images from both \emph{inward} ($\textbf{X}^{in}$) and \emph{outward} ($\textbf{X}^{out}$) directions. Images $\textbf{X}^{in}$ and $\textbf{X}^{out}$ are then fed into the Student network to predict segmentation masks $\widehat{\textbf{Y}}^{in}$ and $\widehat{\textbf{Y}}^{out}$. The segmentation masks are supervised by bidirectional copy-pasting the predictions of the unlabeled images from the Teacher network and the label maps of the labeled images. 


\subsection{Bidirectional Copy-Paste}

\subsubsection{Mean-Teacher and Training Strategy}
In our BCP framework, there are a Teacher network, $\mathcal{F}_{t}\left(\textbf{X}^u_p, \textbf{X}^u_q;\mathbf{\Theta}_{t} \right)$, and a Student network $\mathcal{F}_{s}\left(\textbf{X}^{in}, \textbf{X}^{out};\mathbf{\Theta}_{s} \right)$, where $\mathbf{\Theta}_t$ and $\mathbf{\Theta}_s$ are parameters.
{The Student network is optimized by stochastic gradient descent, and the Teacher network is by exponential moving average (EMA) of Student network.\cite{Tarvainen2017mean}}
Our training strategy is divided into three steps. We first use only labeled data to pretrain a model, then we use the pretrained model as Teacher network to generate pseudo-labels for unlabeled images. At each iteration, we first optimize the Student network parameters $\mathbf{\Theta}_s$ by stochastic gradient descent. Finally, we update the Teacher network parameters $\mathbf{\Theta}_{t}$ using EMA of the Student parameters $\mathbf{\Theta}_{s}$.

\subsubsection{Pre-Training via Copy-Paste}
Inspired by previous work \cite{SimpleCP}, we conducted Copy-Paste augmentation on labeled data to train a supervised model, the supervised model will generate pseudo-labels for unlabeled data during self-training. This strategy was proved to be effective to improve segmentation performance, more details will be illustrated in ablation studies. 

\subsubsection{Bidirectional Copy-Paste Images}
To conduct copy-paste between a pair of images, we first generate a zero-centered mask $\mathcal{M}\in\{0,1\}^{W\times H\times L}$, indicating whether the voxel comes from the foreground ($0$) or the background ($1$) image. The size of the zero-value region is $\beta H\times \beta W\times \beta L$, where $\beta\in(0,1)$. Then we bidirectionally copy-paste labeled and unlabeled images as follows:
\begin{align}
&\textbf{X}^{in}=\textbf{X}^l_{j}\odot\mathcal{M}+\textbf{X}^u_{p}\odot\left(\textbf{1} - \mathcal{M} \right),\label{eq:mix1}\\
&\textbf{X}^{out}=\textbf{X}^u_{q}\odot\mathcal{M}+\textbf{X}^l_{i}\odot\left(\textbf{1} - \mathcal{M} \right),\label{eq:mix2}
\end{align}
where $\textbf{X}^l_{i}$, $\textbf{X}^l_{j}\in\mathcal{D}^l$, $i\neq j$, $\textbf{X}^u_{p}$, $\textbf{X}^u_{q}\in\mathcal{D}^u$, $p\neq q$, $\textbf{1}\in\{1\}^{W\times H\times L}$, and $\odot$ means element-wise multiplication. Two labeled and unlabeled images are adopted to keep the diversity of the input.

\subsubsection{Bidirectional Copy-Paste Supervisory Signals}

To train the Student network, supervisory signals are also generated via BCP operation. Unlabeled images $\textbf{X}^u_p$ and $\textbf{X}^u_q$ are fed into the Teacher network, and their probability maps are computed:
\begin{equation}
    \textbf{P}^u_p=\mathcal{F}_t(\textbf{X}^u_p;\mathbf{\Theta}_t), ~~~\textbf{P}^u_q=\mathcal{F}_t(\textbf{X}^u_q;\mathbf{\Theta}_t).
\end{equation}


The initial pseudo-label $\widehat{\textbf{Y}}^u$ ($p$ and $q$ are dropped for simplicity) are determined by taking a common threshold 0.5 on $\textbf{P}^u$ for binary segmentation tasks, or taking \emph{argmax} operation on $\textbf{P}^u$ for multi-class segmentation tasks. The final pseudo-label $\widetilde{\textbf{Y}}^u$ is obtained via selecting the largest connected component of $\widehat{\textbf{Y}}^u$ which will effectively remove outlier voxels. 
Then, we propose to bidirectionally copy-paste the pseudo-labels of unlabeled images and ground truth labels of labeled images in the same manner as in Eq.\ref{eq:mix1} and Eq.\ref{eq:mix2} to obtain the supervisory signals:

\begin{align}
&\textbf{Y}^{in} = \textbf{Y}^l_{j}\odot\mathcal{M} + \widetilde{\textbf{Y}}^u_{p}\odot\left( \textbf{1} - \mathcal{M} \right),\\
&\textbf{Y}^{out} = \widetilde{\textbf{Y}}^u_{q}\odot\mathcal{M} + \textbf{Y}^l_{i}\odot\left( \textbf{1} - \mathcal{M} \right).
\end{align}
$\textbf{Y}^{in}$ and $\textbf{Y}^{out}$ will be used as the supervision to supervise the Student network predictions of $\textbf{X}^{in}$ and $\textbf{X}^{out}$.

\subsection{Loss Function}
Each input image of the Student network consists of component from both labeled and unlabeled image. Intuitively, ground truth masks of labeled images are usually more accurate than pseudo-labels of unlabeled images. We use $\alpha$ to control the contribution of unlabeled image pixels to the loss function. The loss functions for $\textbf{X}^{in}$ and $\textbf{X}^{out}$ are computed respectively by
\begin{align}
\label{eq:loss}
\mathcal{L}^{in}=\mathcal{L}_\textit{seg}&\left(\textbf{Q}^{in},\textbf{Y}^{in}\right)\odot\mathcal{M}+\\\nonumber
&\alpha\mathcal{L}_\textit{seg}\left(\textbf{Q}^{in},\textbf{Y}^{in}\right)\odot\left(\textbf{1}-\mathcal{M}\right),\\
\mathcal{L}^{out}=\mathcal{L}_\textit{seg}&\left(\textbf{Q}^{out},\textbf{Y}^{out}\right)\odot\left(\textbf{1}-\mathcal{M}\right)+\\\nonumber
&\alpha\mathcal{L}_\textit{seg}\left(\textbf{Q}^{out},\textbf{Y}^{out}\right)\odot\mathcal{M},
\end{align}
where $\mathcal{L}_{seg}$ is the linear combination of Dice loss and Cross-entropy loss. $\textbf{Q}^{in}$ and $\textbf{Q}^{out}$ are computed by:
\begin{equation}
    \textbf{Q}^{in}=\mathcal{F}_s(\textbf{X}^{in};\mathbf{\Theta}_s), ~~~\textbf{Q}^{out}=\mathcal{F}_s(\textbf{X}^{out};\mathbf{\Theta}_s).
\end{equation}
At each iteration we update the parameters $\mathbf{\Theta}_{s}$ in Student network by stochastic gradient descent with the loss function:
\begin{equation}
\mathcal{L}_\textit{all}=\mathcal{L}^{in}+\mathcal{L}^{out}.   
\end{equation}
{Afterwards, Teacher network parameters $\mathbf{\Theta}_{t}^{(k+1)}$ at the $\left( k+1\right)$th iteration are updated:
\begin{equation}
    \mathbf{\Theta}_{t}^{(k+1)} = \lambda \mathbf{\Theta}_{t}^{(k)} + \left( 1 - \lambda \right) \mathbf{\Theta}_{s}^{(k)},
\end{equation}}
where $\lambda$ is the smoothing coefficient parameter.

\subsection{Testing Phase}
In the testing stage, given a testing image $\textbf{X}_{test}$, we obtain the probability map by:
    $\textbf{Q}_{test}=\mathcal{F}(\textbf{X}_{test}; \widehat{\mathbf{\Theta}}_s)$,
where $\widehat{\mathbf{\Theta}}_s$ are the well-trained Student network parameters. The final label map can be easily determined by $\textbf{Q}_{test}$.

\begin{table}[t] \small 
\setlength{\tabcolsep}{0.7mm}{
\resizebox{1\linewidth}{!}{
\begin{tabular}{c|cc|llll}
\hline
\multicolumn{1}{c|}{\multirow{2}{*}{Method}} & \multicolumn{2}{c|}{Scans used}  & \multicolumn{4}{c}{Metrics} \\ 
\cline{2-7} \multicolumn{1}{c|}{} & \multicolumn{1}{l}{Labeled} & \multicolumn{1}{l|}{Unlabeled} & 
Dice$\uparrow$ & Jaccard$\uparrow$ & 95HD$\downarrow$ & ASD$\downarrow$ \\ \hline
\multicolumn{1}{c|}{V-Net} &\multicolumn{1}{c}{4(5\%)} &\multicolumn{1}{c|}{0} &52.55 &39.60 &47.05 &9.87 \\ 
\multicolumn{1}{c|}{V-Net} &\multicolumn{1}{c}{8(10\%)} &\multicolumn{1}{c|}{0} &82.74 &71.72 &13.35 &3.26 \\
\multicolumn{1}{c|}{V-Net} &\multicolumn{1}{c}{80(All)} &\multicolumn{1}{c|}{0} &91.47 &84.36 &5.48 &1.51 \\\hline
UA-MT & \multirow{7}{*}{4(5\%)} & \multirow{7}{*}{76(95\%)} 
& 82.26 & 70.98 & 13.71 & 3.82 \\
SASSNet &  &  & 81.60 & 69.63 & 16.16 & 3.58 \\ 
DTC & & & 81.25 & 69.33 & 14.90 & 3.99 \\
URPC & & & 82.48 & 71.35 & 14.65 & 3.65 \\ 
MC-Net & & & 83.59 & 72.36 & 14.07 & 2.70  \\
SS-Net & & & 86.33 & 76.15 & 9.97 & 2.31 \\ 
Ours & & & \textbf{88.02}{\color{right}\textbf{\scriptsize{$\uparrow$}1.69}}&\textbf{78.72}{\color{right}\textbf{\scriptsize{$\uparrow$}2.57}}&\textbf{7.90}{\color{right}\textbf{\scriptsize{$\downarrow$}2.07}}&\textbf{2.15}{\color{right}\textbf{\scriptsize{$\downarrow$}0.16}} \\ \hline
UA-MT & \multirow{7}{*}{8(10\%)} & \multirow{7}{*}{72(90\%)} 
& 87.79 & 78.39 & 8.68 & 2.12 \\
SASSNet & & & 87.54 & 78.05 & 9.84 & 2.59 \\ 
DTC & & & 87.51 & 78.17 & 8.23 & 2.36  \\
URPC & & & 86.92 & 77.03 & 11.13 & 2.28  \\ 
MC-Net & & & 87.62 & 78.25 & 10.03 & 1.82   \\
SS-Net & & & 88.55 & 79.62 & 7.49 & 1.90 \\
Ours & & & \textbf{89.62}{$\color{right}\textbf{\scriptsize{$\uparrow$}1.07}$}&\textbf{81.31}{\color{right}\textbf{\scriptsize{$\uparrow$}1.69}}&\textbf{6.81}{\color{right}\textbf{\scriptsize{$\downarrow$}0.68}}&\textbf{1.76}{\color{right}\textbf{\scriptsize{$\downarrow$}0.14}}\\ \hline
\end{tabular}}
}
\caption{Comparisons with state-of-the-art semi-supervised segmentation methods on LA dataset. 
Improvements compared with the second best results are {\color{right}\textbf{\scriptsize{highlighted}}}.}
\label{tab:LA}
\end{table}

\section{Experiments}
\subsection{Dataset}

\noindent\textbf{LA dataset}. Atrial Segmentation Challenge \cite{LAdataset} dataset includes 100 3D gadolinium-enhanced magnetic resonance image scans (GE-MRIs) with labels. We strictly follow the setting used in SSNet \cite{SSNet}, DTC \cite{Luo2021DTC} and UA-MT \cite{UAMT}.

\noindent\textbf{Pancreas-NIH}. Pancreas-NIH \cite{NIHPancreas} dataset contains 82 contrast-enhanced abdominal CT volumes which are manually delineated. For fair comparison, we follow the setting in CoraNet \cite{CoraNet}.

\noindent\textbf{ACDC dataset}. ACDC \cite{ACDCdataset} dataset is a four-class (\emph{i.e.} background, right ventricle, left ventricle and myocardium) segmentation dataset, containing 100 patients' scans. The data split \cite{ssl4mis} is fixed with 70, 10, and 20 patients' scans for training, validation, and testing.

\subsection{Evaluation Metrics} We choose four evaluation metrics: \textit{Dice Score} (\%), \textit{Jaccard Score} (\%), \textit{95\% Hausdorff Distance (95HD) in voxel} and \textit{Average Surface Distance (ASD) in voxel}. Given two object regions, Dice and Jaccard mainly compute the percentage of overlap between them, ASD computes the average distance between their boundaries, and 95HD measures the closest point distance between them.

\subsection{Implementation Details}

$\alpha=0.5, \beta=2/3$ are set as the default value in experiments, unless otherwise specified. We conduct all experiments on an NVIDIA 3090 GPU with fixed random seeds.

\begin{table}[t] \small 
\setlength{\tabcolsep}{0.6mm}{
\resizebox{1\linewidth}{!}{
\begin{tabular}{c|cc|llll}
\hline
\multirow{2}{*}{Method} & \multicolumn{2}{c|}{Scans used} &\multicolumn{4}{c}{Metrics} \\ \cline{2-7} & Labeled & Unlabeled & Dice$\uparrow$ & Jaccard$\uparrow$ & 95HD$\downarrow$ & ASD$\downarrow$ \\ \hline
V-Net & \multirow{8}{*}{12(20\%)} & \multirow{8}{*}{50(80\%)} 
& 69.96 & 55.55 & 14.27 & \textbf{1.64}  \\
DAN & & & 76.74 & 63.29 & 11.13 & 2.97  \\
ADVNET & & & 75.31 & 61.73 & 11.72 & 3.88  \\
UA-MT & & & 77.26 & 63.82 & 11.90 & 3.06  \\
SASSNet & & & 77.66 & 64.08 & 10.93 & 3.05 \\
DTC & & & 78.27 & 64.75 & 8.36 & 2.25  \\
CoraNet & & & 79.67 & 66.69 & 7.59 &{1.89}  \\
Ours & & & \textbf{82.91}{$\color{right}\textbf{\scriptsize{$\uparrow$}3.24}$}&\textbf{70.97}{$\color{right}\textbf{\scriptsize{$\uparrow$}4.28}$}&\textbf{6.43}{$\color{right}\textbf{\scriptsize{$\downarrow$}1.16}$}&2.25{$\color{wrong}\textbf{\scriptsize{$\uparrow$}0.61}$}\\\hline
\end{tabular}}
}
\caption{Comparisons with state-of-the-art semi-supervised segmentation methods on the Pancreas-NIH dataset.}
\label{table2}
\end{table}

\begin{table}[t] \small 
\setlength{\tabcolsep}{0.7mm}{
\resizebox{1\linewidth}{!}{
\begin{tabular}{c|cc|llll}
\hline
\multicolumn{1}{c|}{\multirow{2}{*}{Method}} & \multicolumn{2}{c|}{Scans used}  & \multicolumn{4}{c}{Metrics} \\ 
\cline{2-7} \multicolumn{1}{c|}{} & \multicolumn{1}{l}{Labeled} & \multicolumn{1}{l|}{Unlabeled} & 
Dice$\uparrow$ & Jaccard$\uparrow$ & 95HD$\downarrow$ & ASD$\downarrow$ \\ \hline
\multicolumn{1}{c|}{U-Net} &\multicolumn{1}{c}{3(5\%)} &\multicolumn{1}{c|}{0} &47.83 &37.01 &31.16 &12.62 \\ 
\multicolumn{1}{c|}{U-Net} &\multicolumn{1}{c}{7(10\%)} &\multicolumn{1}{c|}{0} &79.41 &68.11 &9.35 &2.70 \\
\multicolumn{1}{c|}{U-Net} &\multicolumn{1}{c}{70(All)} &\multicolumn{1}{c|}{0} &91.44 &84.59 &4.30 &0.99 \\\hline
UA-MT & \multirow{7}{*}{3(5\%)} & \multirow{7}{*}{67(95\%)} 
& 46.04 & 35.97 & 20.08 & 7.75 \\
SASSNet &  &  & 57.77 & 46.14 & 20.05 & 6.06 \\ 
DTC & & & 56.90 & 45.67 & 23.36 & 7.39 \\
URPC & & & 55.87 & 44.64 & 13.60 & 3.74 \\ 
MC-Net & & & 62.85 & 52.29 & 7.62 & 2.33  \\
SS-Net & & & 65.83 & 55.38 & 6.67 & 2.28 \\ 
Ours & & & \textbf{87.59}{$\color{right}\textbf{\scriptsize{$\uparrow$}21.76}$}&\textbf{78.67}{$\color{right}\textbf{\scriptsize{$\uparrow$}23.29}$}&\textbf{1.90}{$\color{right}\textbf{\scriptsize{$\downarrow$}4.77}$}&\textbf{0.67}{$\color{right}\textbf{\scriptsize{$\downarrow$}1.61}$} \\ \hline
UA-MT & \multirow{7}{*}{7(10\%)} & \multirow{7}{*}{63(90\%)} 
& 81.65 & 70.64 & 6.88 & 2.02 \\
SASSNet & & & 84.50 & 74.34 & 5.42 & 1.86 \\ 
DTC & & & 84.29 & 73.92 & 12.81 & 4.01  \\
URPC & & & 83.10 & 72.41 & 4.84 & 1.53  \\ 
MC-Net & & & 86.44 & 77.04 & 5.50 & 1.84   \\
SS-Net & & & 86.78 & 77.67 & 6.07 & 1.40 \\
Ours & & & \textbf{88.84}{$\color{right}\textbf{\scriptsize{$\uparrow$}2.06}$}&\textbf{80.62}{$\color{right}\textbf{\scriptsize{$\uparrow$}2.95}$}&\textbf{3.98}{$\color{right}\textbf{\scriptsize{$\downarrow$}2.09}$}&\textbf{1.17}{$\color{right}\textbf{\scriptsize{$\downarrow$}0.23}$}\\ \hline
\end{tabular}}
}
\caption{Comparisons with state-of-the-art semi-supervised segmentation methods on the ACDC dataset.}
\label{tab:acdc}
\end{table}

\begin{figure}[t]
\includegraphics[width=1\linewidth]{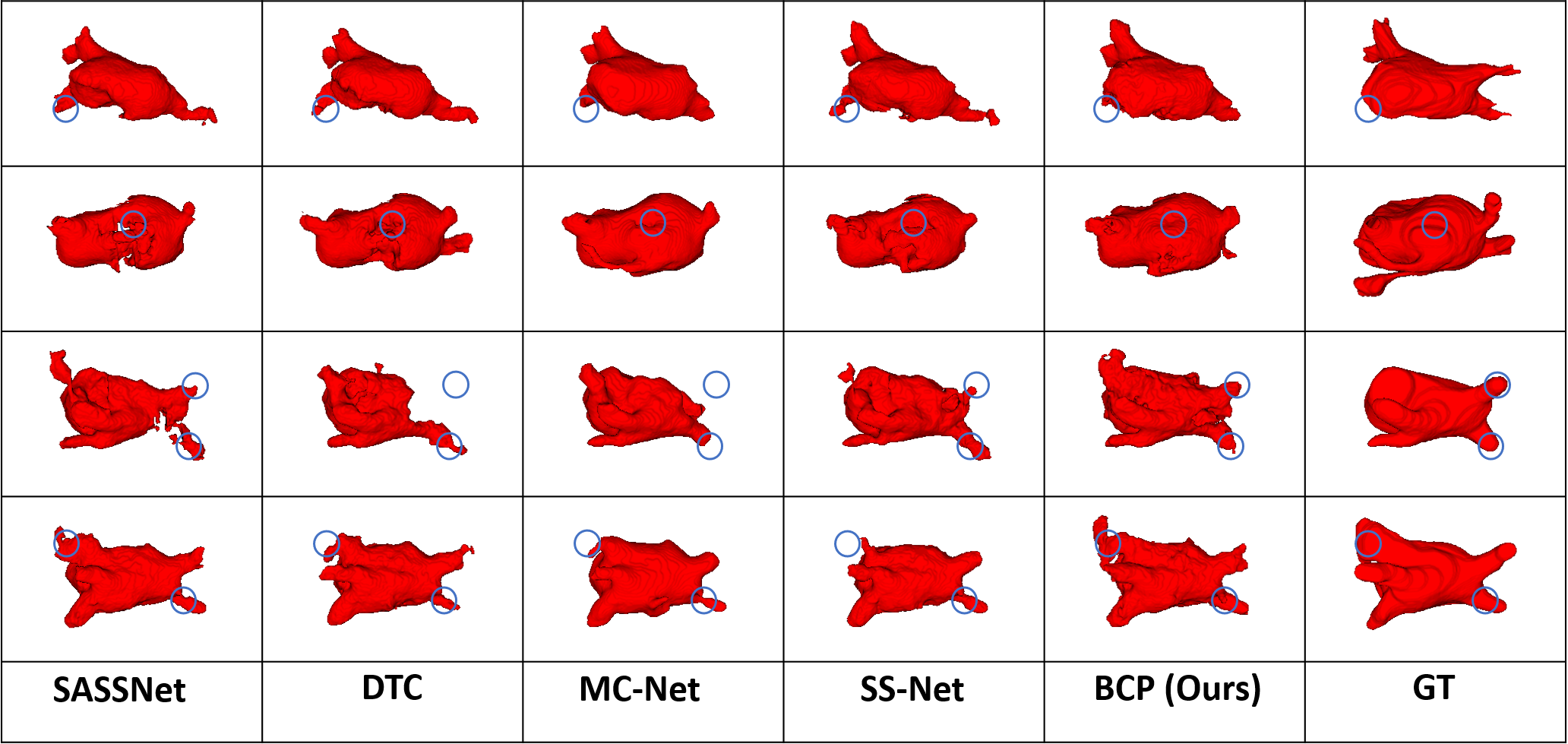}
\centering
\caption{Visualizations of several semi-supervised segmentation methods with 10\% labeled data and ground truth on LA dataset (best viewed by zoom-in on screen).}
\label{fig:visualization}
\end{figure}

\begin{table*}[t]
\centering
\renewcommand\arraystretch{0.93}
\resizebox{0.9\linewidth}{!}{
\begin{tabular}{c|cccccc|cccccc}
\hline
\multirow{3}{*}{Method} & \multicolumn{6}{c|}{LA} & \multicolumn{6}{c}{ACDC}\\ \cline{2-13} 
& \multicolumn{2}{c|}{Scans used}& \multicolumn{4}{c|}{Metrics}
& \multicolumn{2}{c|}{Scans used}&\multicolumn{4}{c}{Metrics} \\ \cline{2-13} 
& Labeled & \multicolumn{1}{c|}{Unlabeled}& Dice$\uparrow$  & Jaccard$\uparrow$  & 95HD$\downarrow$  & ASD$\downarrow$  
& Labeled & \multicolumn{1}{c|}{Unlabeled}& Dice$\uparrow$  & Jaccard$\uparrow$  & 95HD$\downarrow$  & ASD$\downarrow$  \\ \hline
In & \multirow{4}{*}{4(5\%)} & \multicolumn{1}{c|}{\multirow{4}{*}{76(95\%)}} 
& 87.35 & 77.77   & 8.75  & 2.21 & \multirow{4}{*}{3(5\%)}  & \multicolumn{1}{c|}{\multirow{4}{*}{67(95\%)}} & 81.68 & 70.07   & 4.69  & 1.28  \\
Out& & \multicolumn{1}{c|}{}& 87.32 & 77.78 & 9.38  & 2.16 & &\multicolumn{1}{c|}{} 
& 72.19 & 60.69 & 39.57 & 18.15 \\
CP && \multicolumn{1}{c|}{} & 79.67 & 67.05 & 14.66 & 3.21 & & \multicolumn{1}{c|}{} 
& 81.80 & 71.70 & 16.29 & 6.43  \\
Ours &  & \multicolumn{1}{c|}{} & \textbf{88.02} & \textbf{78.72} & \textbf{7.90}  & \textbf{2.15} & & \multicolumn{1}{c|}{}
& \textbf{87.59} & \textbf{78.67} & \textbf{1.90}   & \textbf{0.67}  \\ \hline
In & \multirow{4}{*}{8(10\%)}& \multicolumn{1}{c|}{\multirow{4}{*}{72(90\%)}} 
& 89.02 & 80.38 & 8.08  & 1.81 & \multirow{4}{*}{7(10\%)} & \multicolumn{1}{c|}{\multirow{4}{*}{63(90\%)}} & 85.55 & 75.65   & 4.93  & 1.50  \\
Out & & \multicolumn{1}{c|}{}& 87.61 & 78.10 &8.99 & 2.63 &  & \multicolumn{1}{c|}{}   & 87.23 & 78.07   & 8.61  & 2.39  \\
CP & & \multicolumn{1}{c|}{}& 86.74 & 77.18   & 8.65  & 2.26 & & \multicolumn{1}{c|}{}   
& 88.17 & 79.64   & 6.14  & 1.45  \\
Ours& & \multicolumn{1}{c|}{}& \textbf{89.62} & \textbf{81.31}   & \textbf{6.81}  & \textbf{1.76} & & \multicolumn{1}{c|}{} & \textbf{88.84} & \textbf{80.62} & \textbf{3.98}  & \textbf{1.17}  \\ \hline
\end{tabular}
}
\caption{Ablation study of the copy-paste directions. In: \emph{inward} copy-paste (foreground: unlabeled, background: labeled). Out: \emph{outward} copy-paste (foreground: labeled, background: unlabeled). CP: \emph{direct} copy-paste (background \& foreground: labeled \& labeled and unlabeled \& unlabeled).}
\label{tab:CPdirection}
\vspace{-0.8em}
\end{table*}

\begin{figure}[t]
\includegraphics[width=1\linewidth]{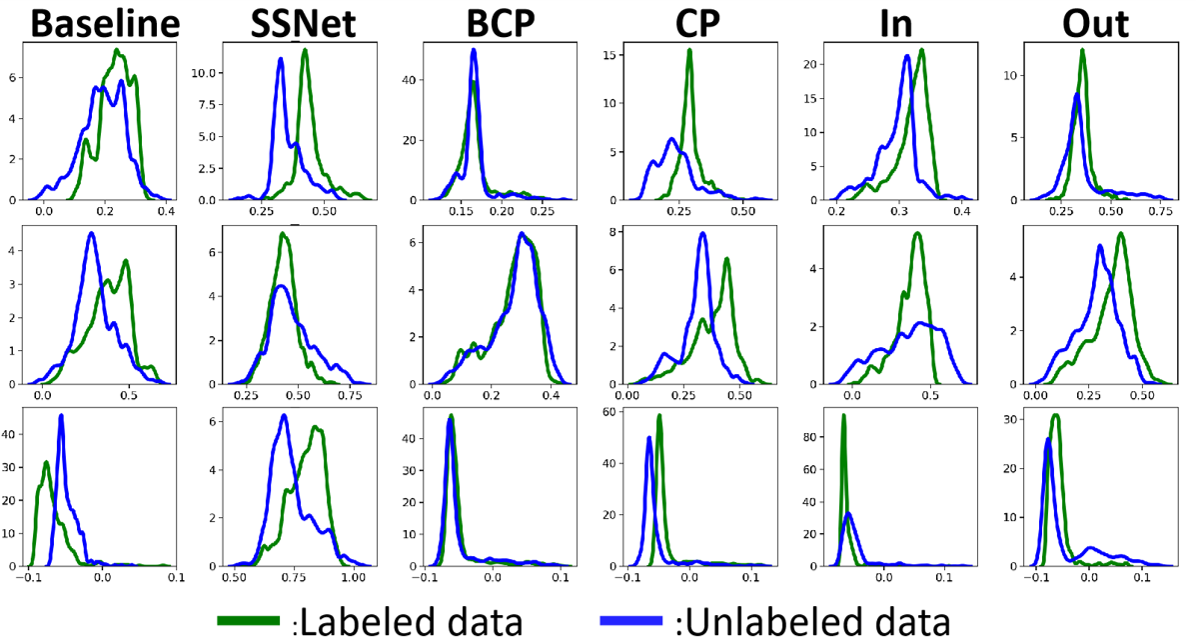}
\centering
\caption{{Kernel dense estimations of different methods, trained on 10\% labeled ACDC dataset. Top to bottom are kernel dense estimations of features belong to three different class of ACDC: right ventricle, myocardium and left ventricle. Baseline: Only labeled data are used to train the network. \emph{CP}, \emph{In} and \emph{Out} are same as Table \ref{tab:CPdirection}. It can be seen that our BCP could make the features of labeled data and unlabeled data align better. Furthermore, the outstanding performance of our method compared with \emph{In} and \emph{Out} demonstrates the necessity of \emph{bidirectional} copy-paste}. }
\label{fig:kde_3class}
\vspace{-0.9em}
\end{figure}

\begin{table}[t]
\resizebox{1\linewidth}{!}{
\begin{tabular}{c|cc|cccc}
\hline
\multirow{2}{*}{Method} & \multicolumn{2}{c|}{Scans used}                      & \multicolumn{4}{c}{Metrics}     \\ \cline{2-7} 
                        & Labeled                  & Unlabeled                 & Dice$\uparrow$  & Jaccard$\uparrow$ & 95HD$\downarrow$  & ASD$\downarrow$   \\ \hline
Mixup                  & \multirow{3}{*}{4(5\%)}  & \multirow{3}{*}{76(95\%)} & 41.71 & 29.58   & 59.75 & 21.87 \\
FG-CutMix                   &                          &                           & 67.05 & 54.00   & 30.52 & 6.15  \\
Ours                    &                          &                           & \textbf{88.02} & \textbf{78.72}   & \textbf{7.90}  & \textbf{2.15}  \\ \hline
Mixup                  & \multirow{3}{*}{8(10\%)} & \multirow{3}{*}{72(90\%)} & 63.64 & 52.51   & 21.67 & 3.61  \\
FG-CutMix                    &                          &                           & 83.58 & 72.70   & 11.96 & 2.56  \\
Ours                    &                          &                           & \textbf{89.62} & \textbf{81.31}   & \textbf{6.81}  & \textbf{1.76}  \\ \hline
\end{tabular}}
\caption{Ablation study of interpolation strategies on LA dataset. Mixup: We imitate the framework of GuidedMix-Net\cite{GuidedMix-Net}, which is proposed for semi-supervised segmentation of natural images. {FG-CutMix}: We crop images of the whole training batch into 4$\times$4 patches and then combine them randomly to generate new images.}
\label{tab:Interpolation}
\vspace{-0.8em}
\end{table}

\noindent\textbf{LA dataset}. Following SS-Net \cite{SSNet}, we use rotation and flip operations to augment data and train our model via an SGD optimizer with the initial learning rate 0.01 decay by 10\% every 2.5K iterations. The backbone is set as 3D V-Net. During training, we randomly crop $112\times112\times80$ patches, and the size of the zero-value region is $74\times74\times53$ ($\beta=2/3$). The batch size is set as 8, containing four labeled patches and four unlabeled patches. The iterations of pre-training and self-training are set as 2k and 15k respectively. 

\noindent\textbf{Pancreas-NIH}. Following CoraNet \cite{CoraNet}, we augment data by rotating, rescaling and flipping, and train a four-layer 3D V-Net by Adam optimizer with initial learning rate as 0.001. During training, we randomly crop $96\times96\times96$ patches input the network, the size of the zero-value region of mask $\mathcal{M}$ is $64\times64\times64$. We set the batch size, pre-training epochs and the self-training epochs as 8, 60 and 200 respectively. 

\noindent\textbf{ACDC dataset}. Following SS-Net \cite{SSNet}, we use 2D U-Net as the backbone of our experiments. During training, the input patch size is $256\times256$ (2D slices) and the size of the zero-value region of mask $\mathcal{M}$ is $170\times170$. The batch size, pre-training iterations, and the self-training training iterations are set as 24, 10k and 30k respectively. 

\subsection{Comparison with Sate-of-the-Art Methods}

\paragraph{LA dataset}
We compare our framework on LA dataset with various competitors: UA-MT \cite{UAMT}, SASSNet \cite{SASSNet}, DTC \cite{Luo2021DTC}, URPC \cite{URPC}, MC-Net \cite{MCNet} and SS-Net \cite{SSNet}. Following SS-Net, semi-supervised experiments of different labeled ratios (\emph{i.e.} 5\% and 10\%) are carried out. Results from other competitors were reported in the identical experimental setting in SS-Net \cite{SSNet} for fair comparisons. As shown in Table~\ref{tab:LA}, our method achieves the best performance on all four evaluation metrics, outperforming other competitors by a big margin. 
Thanks to BCP, the network ``sees" more variances for boundary regions or semantic change of voxels, allowing for achieving good shape-related performances (see 95HD and ASD) without any explicit boundary or shape constraints during training. 
Moreover, it can be seen in Fig. \ref{fig:visualization} that our method can segment fine details of the target organ, especially edges that are easy to be misidentified (the first row) or missed (the second, third and fourth row), highlighted by blue circles. 

\paragraph{Pancreas-NIH dataset}
We conduct experiments on Pancreas-NIH dataset with 20\% labeled ratio \cite{Luo2021DTC, CoraNet}. We compared BCP with V-Net \cite{VNet}, DAN \cite{DAN}, ADVNET \cite{ADVNet}, UA-MT \cite{UAMT}, SASSNet \cite{SASSNet}, DTC \cite{Luo2021DTC} and CoraNet \cite{CoraNet} in Table \ref{table2}. In this table, DAN, ADVNET, UA-MT, SASSNet, DTC, CoraNet and our method took both labeled and unlabeled data to train the network with V-Net as the backbone, while V-Net only uses labeled data in the supervised setting (lower bound). BCP achieves significant improvement on Dice, Jaccard and 95HD (\emph{i.e.}, surpassing the second best by 3.24\%, 4.28\% and 1.16, respectively). These results do not conduct any post-processing for fair comparison\vspace{-0.5em}.

\begin{table}[t]\small
\renewcommand\arraystretch{0.95}
\resizebox{1\linewidth}{!}{
\begin{tabular}{c|cc|cccc}
\hline
\multirow{2}{*}{Mode} & \multicolumn{2}{c|}{Scans used}                  & \multicolumn{4}{c}{Metrics} \\ \cline{2-7} 
                            & Labeled                & Unlabeled    & Dice$\uparrow$  & Jaccard$\uparrow$  & 95HD$\downarrow$ & ASD$\downarrow$ \\ \hline
Random                      & \multirow{3}{*}{4(5\%)}  & \multirow{3}{*}{76(95\%)} & 86.15    & 76.03       & 9.19    & 2.38   \\
Contact                     &                        &                         & 86.64    & 76.32       & 9.61    & 2.58   \\
Ours                        &                        &                         & \textbf{88.02} & \textbf{78.72}&\textbf{7.90}&\textbf{2.15}   \\ \hline
Random                      & \multirow{3}{*}{8(10\%)} & \multirow{3}{*}{72(90\%)} & 84.50    & 73.79       & 10.79    & 2.39   \\
Contact                     &                        &                         & 88.66    & 79.82       & 7.93    & 2.27   \\
Ours                        &                        &                         & \textbf{89.62} &\textbf{81.31} &\textbf{6.81} & \textbf{1.76}   \\ \hline
\end{tabular}
}
\caption{Results with three masking strategies on LA dataset.}
\label{tab:masking}
\end{table}

\paragraph{ACDC dataset}
Table \ref{tab:acdc} shows the averaged performance of four-class segmentation results on ACDC dataset with 5\% and 10\% labeled ratios. BCP surpasses all state-of-the-arts. We obtain a huge performance improvement up to 21.76\% in terms of Dice for the setting with 5\% labeled ratio. Following SS-Net \cite{SSNet}, 2D slices are used to train our network. Noted that one 3D volume can be sliced into many 2D slices, so much more combinations from labeled and unlabeled slices could be produced than those using 3D data. Hence, during training, the knowledge of labeled data can be transferred to unlabeled data more sufficiently, especially when the number of labeled volume is very small. This might be the reason for such a significant improvement compared with others when the labeled ratio is 5\%. 

\subsection{Ablation Studies} 
We conduct ablation studies to show the impact of each component in BCP. Including CP directions, design choices of masking strategies, {{ interpolation strategies}}, $\beta$ (size ratio for zero-value region in $\mathcal{M}$), and $\alpha$ (Eq.~\ref{eq:loss}-{\color{red}7}). We also investigate step-by-step the significant improvement of our method compared with competitors on the ACDC dataset with 5\% labeled ratio. {{Some ablation studies on ACDC dataset are shown in the supplementary material\vspace{-0.5em}.}}

\begin{table}[t]\small
\renewcommand\arraystretch{0.93}
\resizebox{0.95\linewidth}{!}
{
\begin{tabular}{c|cc|cccc}
\hline
\multirow{2}{*}{$\beta$} & \multicolumn{2}{c|}{Scans used}                  & \multicolumn{4}{c}{Metrics} \\ \cline{2-7} 
                            & Labeled                & Unlabeled  & Dice$\uparrow$  & Jaccard$\uparrow$  & 95HD$\downarrow$ & ASD$\downarrow$ \\ \hline
1/3                      & \multirow{3}{*}{4(5\%)}  & \multirow{3}{*}{76(95\%)} & 79.92    & 67.73       & 15.44    & 3.63   \\
1/2                     &                        &                         & 86.49    & 76.63       & 8.74    & 2.23   \\
2/3                        &                        &                         & \textbf{88.02} & \textbf{78.72}&\textbf{7.90}&\textbf{2.15}   \\ 
5/6                     &                        &                         & 87.92    & 78.57       & 8.29    & 2.26   \\\hline
1/3                      & \multirow{3}{*}{8(10\%)} & \multirow{3}{*}{72(90\%)} & 83.20    & 72.04       & 11.64    & 2.94   \\
1/2                     &                        &                         & 88.81    & 89.10       & 7.33    & 1.96   \\
2/3                        &                        &                         & \textbf{89.62} &\textbf{81.31} &\textbf{6.81} & \textbf{1.76}   \\ 
5/6                     &                        &                         & 88.75    & 79.96       & 7.63   & 2.07   \\\hline
\end{tabular}
}
\caption{Ablation study of $\beta$ on LA dataset.}
\label{table7}
\vspace{-0.8em}
\end{table}

\begin{table*}[t]
\centering
\renewcommand\arraystretch{0.95}
\resizebox{0.8\linewidth}{!}{
\begin{tabular}{c|cccccc|cccccc}
\hline
\multirow{3}{*}{$\alpha$} & \multicolumn{6}{c|}{LA} & \multicolumn{6}{c}{ACDC}\\ \cline{2-13} 
& \multicolumn{2}{c|}{Scans used}& \multicolumn{4}{c|}{Metrics}
& \multicolumn{2}{c|}{Scans used}&\multicolumn{4}{c}{Metrics} \\ \cline{2-13} 
& Labeled & \multicolumn{1}{c|}{Unlabeled}& Dice$\uparrow$  & Jaccard$\uparrow$  & 95HD$\downarrow$  & ASD$\downarrow$  
& Labeled & \multicolumn{1}{c|}{Unlabeled}& Dice$\uparrow$  & Jaccard$\uparrow$  & 95HD$\downarrow$  & ASD$\downarrow$  \\ \hline
0.5 & \multirow{4}{*}{4(5\%)} & \multicolumn{1}{c|}{\multirow{4}{*}{76(95\%)}} 
& \textbf{88.02} & \textbf{78.72} & \textbf{7.90}  & \textbf{2.15} & \multirow{4}{*}{3(5\%)}  & \multicolumn{1}{c|}{\multirow{4}{*}{67(95\%)}} & \textbf{87.59} & \textbf{78.67} & \textbf{1.90} & \textbf{0.67}  \\
1.5& & \multicolumn{1}{c|}{}& 87.21 & 77.49 &8.67 & 2.37 & &\multicolumn{1}{c|}{} 
& 85.88 & 76.02 &3.17 & 0.93   \\
2.5 &  & \multicolumn{1}{c|}{} & 86.56 & 76.46 & 9.82 & 2.60 & & \multicolumn{1}{c|}{}
& 85.43 & 75.47 & 12.02 & 4.05  \\ \hline
0.5 & \multirow{4}{*}{8(10\%)}& \multicolumn{1}{c|}{\multirow{4}{*}{72(90\%)}} 
& \textbf{89.62} & \textbf{81.31} & \textbf{6.81}  & \textbf{1.76} & \multirow{4}{*}{7(10\%)} & \multicolumn{1}{c|}{\multirow{4}{*}{63(90\%)}} & \textbf{88.84} & \textbf{80.62} & 3.98  & 1.17  \\
1.5 & & \multicolumn{1}{c|}{}& 89.35 & 80.88 & 7.46 & 2.09 &  & \multicolumn{1}{c|}{} & 88.65 & 80.31   & \textbf{1.99}  & \textbf{0.68}  \\
2.5& & \multicolumn{1}{c|}{}& 88.74 & 79.88   & 7.73  & 2.15 & & \multicolumn{1}{c|}{} & 87.13 & 78.19 & 3.67 & 1.24  \\ \hline
\end{tabular}
}
\caption{Ablation study of the weights $\alpha$ in the loss function.}
\label{table5}
\vspace{-0.8em}
\end{table*}

\begin{figure}[t]
\includegraphics[width=0.6\columnwidth]{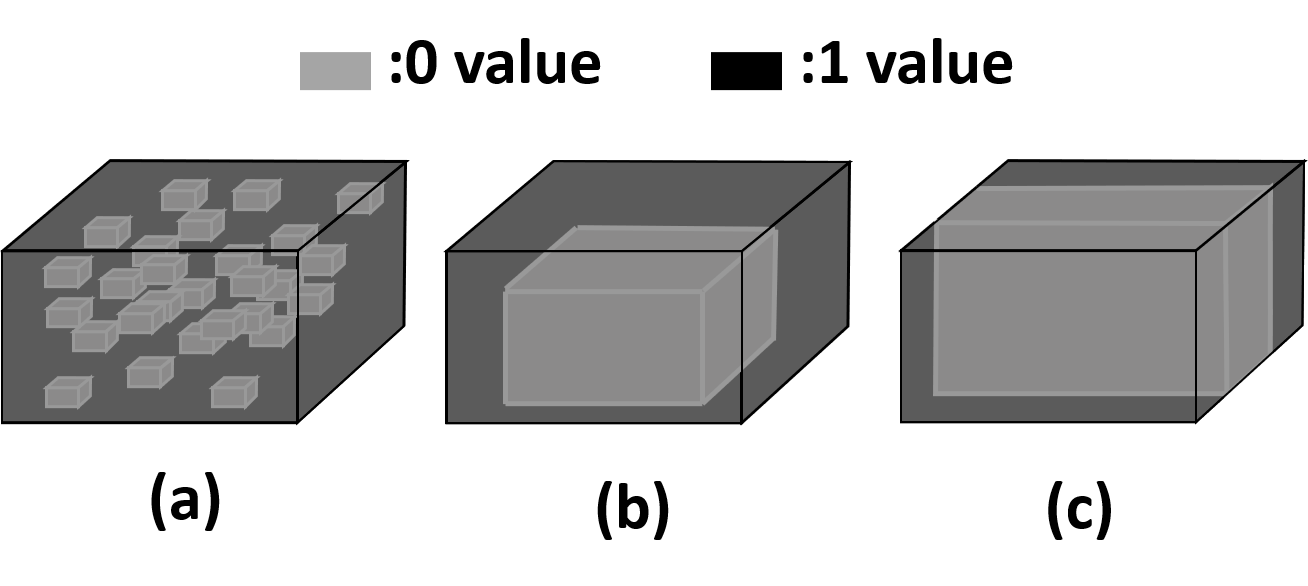}
\centering
\vspace{-0.5em}
\caption{Different masking strategies. (a): random mask; (b): zero-centered mask; (c): contact mask.}
\label{fig:masking}
\vspace{-0.8em}
\end{figure}

\paragraph{Copy-Paste Direction}
We design three experiments to investigate the influence of different copy-paste directions in Table \ref{tab:CPdirection}. \emph{Inward} and \emph{outward} copy-paste (\textit{In} and \textit{Out} in the table) mean using $\textbf{X}^l\odot\mathcal{M}+\textbf{X}^u\odot\left(\textbf{1} - \mathcal{M} \right)$ or $\textbf{X}^u\odot\mathcal{M}+\textbf{X}^l\odot\left(\textbf{1} - \mathcal{M} \right)$ respectively to train the network. We also conduct within-set copy-paste (\textit{CP} in the table), \emph{i.e.}, copy-paste labeled data on another labeled data and copy-paste unlabeled data on another unlabeled data. We can see that all these variants get inferior performances compared with our BCP, since they either lack consistent manner for training labeled and unlabeled data, or lack common semantics transfer between labeled and unlabeled data. 

\paragraph{Interpolation Strategies}
We compare BCP with other two interpolation strategies: Mixup \cite{Mixup} and Fine-Grained CutMix (FG-CutMix).
{For Mixup, we superimpose labeled and unlabeled data to generate new training images, imitating the framework of GuidedMix-Net \cite{GuidedMix-Net}. For FG-CutMix, we crop training images into 4$\times$4 patches and combine them in batch to generate new images.}
The network predictions of new images are re-combined and then supervised by ground-truth or pseudo-labels. The results of LA dataset are shown in Table \ref{tab:Interpolation}. Due to similar spatial structures of medical images, Mixup brings more influential noise in medical images. FG-CutMix maintains less structure information after CutMix than BCP. More details will be discussed in supplementary material\vspace{-0.5em}.

\begin{table}[t]\small
\resizebox{1\linewidth}{!}{
\begin{tabular}{c|cc|cccc}
\hline
\multirow{2}{*}{Strategy} & \multicolumn{2}{c|}{Scans used}                  & \multicolumn{4}{c}{Metrics} \\ \cline{2-7} 
                            & Labeled                & Unlabeled    & Dice$\uparrow$  & Jaccard$\uparrow$  & 95HD$\downarrow$ & ASD$\downarrow$ \\ \hline
random                      & \multirow{3}{*}{4(5\%)}  & \multirow{3}{*}{76(95\%)} & 86.06    & 75.96       & 9.48    & 2.33   \\
w/o CP                    &                        &                         & 86.46    & 76.50       & 8.93    & 2.31   \\
Ours                        &                        &                         & \textbf{88.02} & \textbf{78.72}&\textbf{7.90}&\textbf{2.15}   \\ \hline
random                      & \multirow{3}{*}{8(10\%)} & \multirow{3}{*}{72(90\%)} & 87.93    & 78.67       & 8.24    & 2.08   \\
w/o CP                     &                        &                         & 88.75    & 79.88       & 7.66    & 1.83   \\
Ours                        &                        &                         & \textbf{89.62} &\textbf{81.31} &\textbf{6.81} & \textbf{1.76}   \\ \hline
\end{tabular}}
\caption{Ablation study of pre-training strategy on LA dataset. random: Initialized randomly. w/o CP: Initialized from a pre-trained model trained on labeled data without copy-paste.}
\label{table8}
\vspace{-1em}
\end{table}

\paragraph{Design Choices of Masking Strategies}
As shown in Fig.~\ref{fig:masking}, we explore different masking strategies in BCP on LA dataset. To conduct a fair comparison, we maintain the same number of zero-value voxels for different strategies. We randomly sample 27 small $\beta H\times \beta W\times \beta L$ zero-value cubes in an all-one mask, and set $\beta=2/9$ for each zero-value cube. For contact mask, the shape of zero-value region is $ \beta H\times W\times L$, and $\beta$ is set as 8/27 to control the number of zero-value voxels. As shown in Table \ref{tab:masking}, random mask obtains worst performance, since small random cubes only contain incoherently local foreground of an image, which lacks the ability in learning complete foreground representation.  
Contact mask has better integrity of foreground information, which performs better than random mask, but still performs worse than zero-centered mask used in our method, since foreground has less chance interacting with the background, compared with zero-centered mask. Thus, both random mask and contact mask have weaker ability in mitigating the distribution mismatch problem between labeled and unlabeled data\vspace{-0.5em}.

\paragraph{Size of Zero-value Region in M}
We study the impact of zero-value region size on LA dataset, as shown in the Table \ref{table7}. For the zero-value region $\beta H\times \beta W\times \beta L$ in the mask $\mathcal{M}$, we set $\beta=\{\frac{1}{3}, \frac{1}{2}, \frac{2}{3}, \frac{5}{6}\}$. The performance gets worse as $\beta$ decreases, which means small copy-pasted foreground has limited ability in transferring common semantics to/from the background. Best performance is achieved when $\beta=2/3$ and it decreases a bit when $\beta= 5/6$\vspace{-0.8em}. 

\paragraph{Weight in Loss Function}
We set $\alpha=0.5$ as the default value. Now we vary $\alpha=\{0.5,1.5,2.5\}$ to see how the performance changes. Table \ref{table5} shows it is not sensitive when $\alpha$ changes from 0.5 to 1.5, but an obvious performance drop is observed when $\alpha=2.5$\vspace{-0.8em}. 

\begin{table}[t]
\renewcommand\arraystretch{0.95}
\resizebox{0.9\linewidth}{!}{
\begin{tabular}{ccc|cccc}
\hline
BCP & nms & Pre-Train & Dice$\uparrow$ & Jaccard$\uparrow$ & 95HD$\downarrow$ & ASD$\downarrow$ \\ \hline
    &     &          & 47.62    & 36.61       & 29.02    & 11.46   \\
\checkmark   &     &          & 83.26    & 72.71       & 23.90    & 7.49   \\
\checkmark  & \checkmark   &          & 82.33    & 72.76       & 9.78    & 4.74   \\
\checkmark   & \checkmark   & \checkmark    & \textbf{87.59}& \textbf{78.67}& \textbf{1.90}& \textbf{0.67}   \\ \hline
\end{tabular}
}
\caption{Ablation on ACDC dataset with 5\% labeled data, $\alpha=0.5$ across all experiments. nms: Post-processing the pseudo-labels for unlabeled data. Pre-Train: Initialized from a pre-trained model with copy-paste on labeled data.}
\label{table9}
\vspace{-0.8em}
\end{table}

\paragraph{Teacher Network Initialization Strategy}
In our default setting, the Teacher network is initialized by a pre-trained model, which is trained on labeled data in a copy-paste manner. We study other network initialization strategies: initialized randomly and initialized from a pre-trained model which is trained on labeled data without copy-paste. Comparison results are shown in Table \ref{table8}. Compared with pre-training on original labeled data, performing copy-paste on the labeled data during pre-train can effectively improve the generalization ability of the network\vspace{-0.8em}.

\paragraph{Ablation on ACDC with 5\% Labeled Data}
In Table \ref{tab:acdc}, our method achieves a huge improvement on ACDC dataset with 5\% labeled data. We separate our method into three components and study which component contributes the most to this improvement. As shown in Table \ref{table9}, without our component (the first row), it degenerated into a normal pseudo-label-based self-training method, which means the segmentation of labeled and unlabeled images are supervised by ground truth and pseudo-labels respectively. Then, BCP leads to a significant performance gain (from 47.62\% to 83.26\% in Dice). Post-processing (nms) and a better Teacher network initialization enhance the quality of pseudo-labels and thus further improve the performance.

\section{Conclusion}
We have presented the bidirectional copy-paste (BCP) for semi-supervised medical image segmentation. We extend copy-paste-based method in a bidirectional manner, which reduces the distribution gap between labeled and unlabeled data. Experiments on LA, NIH-Pancreas and ACDC datasets show the superiority of the proposed BCP, with even over 21\% Dice improvement on ACDC dataset with 5\% labeled data. Note that BCP does not introduce new parameters or computational cost compared with the backbone network. \noindent\textbf{Limitations.} We didn't specifically design a module to enhance local attributes learning. Though BCP performs better than all competitors, target parts with extremely low contrast are still hard to segment well (\emph{e.g.}, bottom left part on
$2$nd row of Fig. 4 is missing).
\section{Acknowledgement}
This work was supported by the National Natural Science Foundation of China (Grant No. 62101191), Shanghai Natural Science Foundation (Grant No. 21ZR1420800), the Science and Technology Commission of Shanghai Municipality (Grant No. 22DZ2229004), and the Fundamental Research Funds for the Central Universities.

{\small
\bibliographystyle{ieee_fullname}
\bibliography{egbib}
}

\end{document}